\documentclass[10pt,twocolumn,letterpaper]{article}

\usepackage{cvpr}              
\definecolor{cvprblue}{rgb}{0.21,0.49,0.74}
\usepackage[pagebackref,breaklinks,colorlinks,allcolors=cvprblue]{hyperref}
\usepackage{multirow}
\usepackage{algorithm}
\usepackage{algorithmic}
\usepackage{float}
\usepackage{stfloats}
\def\paperID{32884} 
\def\confName{CVPR}
\def\confYear{2026}

\title{Locate-then-Sparsify: Attribution Guided Sparse Strategy \\ for Visual Hallucination Mitigation}

\author{Tiantian Dang$^{1,2}$\quad
Chao Bi$^{1}$\thanks{Corresponding author.}\quad
Shufan Shen$^{1}$\quad
Jinzhe Liu$^{1}$\\
Qingming Huang$^{1,3}$\quad
Shuhui Wang$^{1,2*}$
\\
$^{1}$State Key Lab. of AI Safety, Institute of Computing Technology, Chinese Academy of Sciences\\
$^{2}$School of Advanced Interdisciplinary Sciences, University of Chinese Academy of Sciences\\
$^{3}$ School of Computer Science and Technology, University of Chinese Academy of Sciences\\
{\tt\small dangtiantian23@mails.ucas.ac.cn}\quad
{\tt\small qmhuang@ucas.ac.cn}\\
{\tt\small \{bichao,shenshufan22z,liujinzhe23b,wangshuhui\}@ict.ac.cn}
}

\begin{document}
\maketitle
\begin{abstract}
Despite the significant advancements in Large Vision-Language Models~(LVLMs), their tendency to generate hallucinations undermines reliability and restricts broader practical deployment.
Among the hallucination mitigation methods, feature steering emerges as a promising approach that reduces erroneous outputs in LVLMs without increasing inference costs.
However, current methods apply uniform feature steering across all layers. This heuristic strategy ignores inter-layer differences, potentially disrupting layers unrelated to hallucinations and ultimately leading to performance degradation on general tasks.
In this paper, we propose \textbf{L}ocate-\textbf{T}hen-\textbf{S}parsify for \textbf{F}eature \textbf{S}teering (\textbf{LTS-FS}), a plug-and-play framework which controls the steering intensity according to the hallucination relevance of each layer.
We first construct a dataset comprising token-level and sentence-level hallucination cases. Based on this dataset, we introduce an attribution method based on causal interventions to quantify the hallucination relevance of each layer. 
With the attribution scores across layers, we propose a layerwise strategy that converts these scores into feature steering intensities for individual layers, enabling more precise adjustments specifically on hallucination-relevant layers.
Extensive experiments across multiple LVLMs and benchmarks demonstrate that LTS-FS effectively mitigates hallucination while preserving strong performance. Codes are available at \url{https://github.com/huttersadan/LTS-FS}.
\end{abstract}

\vspace{-2mm}

\section{Introduction}
\label{sec:intro}

By harnessing the advanced text generation capabilities of Large Language Models, Large Vision Language Models~(LVLMs) have achieved impressive performance across various multimodal tasks~\cite{flamingo_2022,liu2023visual,qwen2vl_2024,lvml_survey_2024}.
Despite their strong performance, LVLMs face a significant challenge known as \textit{hallucination}, wherein the model generates fluent and semantically coherent responses that include factually incorrect statements about the input visual content~\cite{li2023evaluating,lvml_hallu_eval_2024,lvml_survey_2024}.
Such hallucinations hinder the reliability of LVLMs, posing serious risks in real-world applications~\cite{hallucination_survey_2023,mitigation_survey_2024}.

\begin{figure}[t]
    \centering
    \begin{subfigure}[t]{0.88\linewidth}
        \centering
        \includegraphics[width=\linewidth]{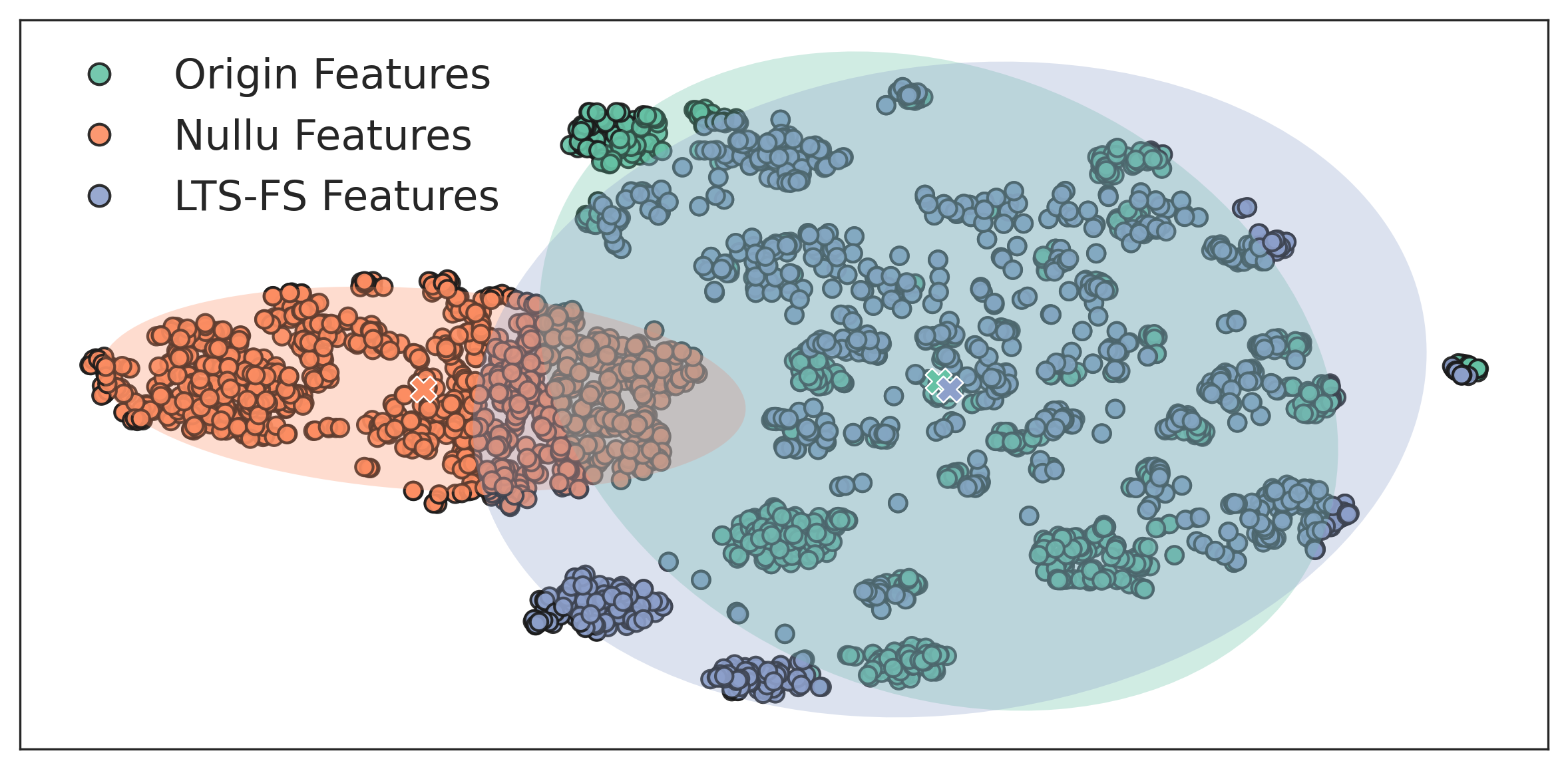}
        \caption{TSNE visualizations of features in LVLM layers.}
        \label{fig:tsne}
    \end{subfigure}
    \hfill
    \begin{subfigure}[t]{1.0\linewidth}
        \centering
        \includegraphics[width=\linewidth]{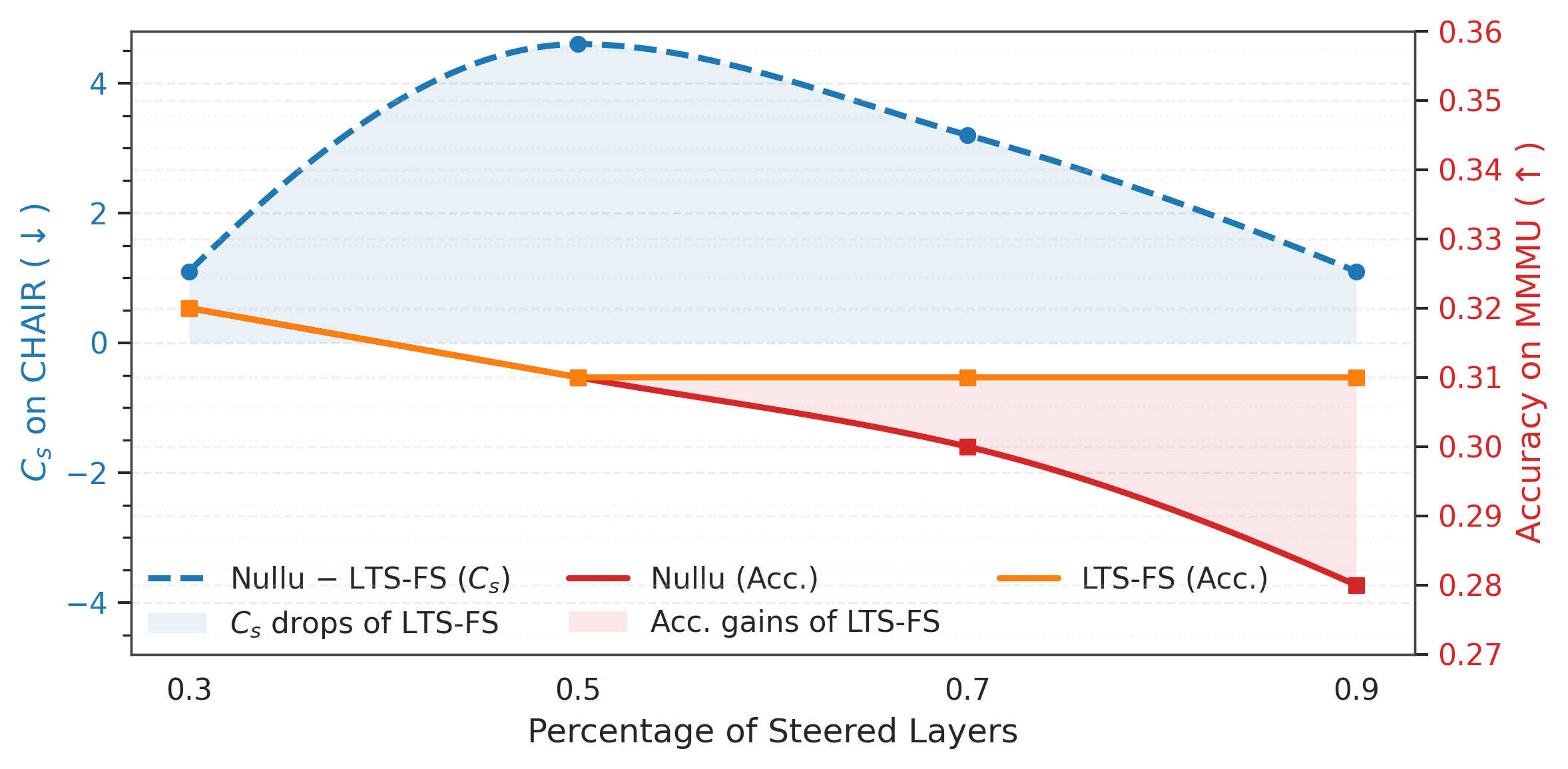}
        \caption{Performance on CHAIR and MMMU benchmarks.}
        \label{fig:intro-mmmu}
    \end{subfigure}
   
    \caption{Current methods~(\textit{e.g.}, Nullu~\cite{yang2025nullu}) mitigate hallucinations by uniformly steering features across layers, which~(a)~alters feature distributions and~(b)~ In contrast, we propose a layerwise steering framework LTS-FS, which mitigates hallucinations more effectively~(\textit{e.g.}, on CHAIR) while minimally perturbing the feature distributions, preserving more generalization ability.}
    \label{fig:intro}
    \vspace{-4ex}
\end{figure}

To mitigate hallucinations in LVLMs, early studies finetune the whole model on specific datasets, which is costly and leads to degradation of generalization ability~\cite{liu2023mitigating, wang2024vigc,8957311}. In contrast, decoding-based methods introduce strategies such as contrastive decoding~\cite{leng2024mitigating,An_2025_CVPR} and self-correction~\cite{yin2024woodpecker,bi2025asking} in a training-free manner.
Nevertheless, these methods require many more decoding steps for each input query, leading to high inference costs in real-world deployment.
Recently, feature steering methods~\cite{yang2025nullu,liu2025reducing} show advantages in overcoming the above limitations. 
These methods steer features of intermediate layers toward directions that are less prone to generating hallucinated outputs. By modifying only the features without introducing additional decoding steps, feature steering methods can maintain inference costs comparable to those of the original model.
However, current methods steer features based on heuristically designed rules~\cite{liu2025reducing},~\textit{e.g.}, adjusting all layers. 
These rules overlook the inherent differences across layers in pre-trained models, making the steering process work on layers less relevant to hallucinations. The disruption alters the distributions of features (see~\cref{fig:tsne}) and ultimately impairs the model's generalization ability~(see~\cref{fig:intro-mmmu}), similar to the tuning-based methods.
Therefore, methods that can steer features to the right location and preserve the original capabilities of LVLMs are urgently required.

\begin{figure*}[t]
    \centering
    \includegraphics[width=0.97\linewidth]{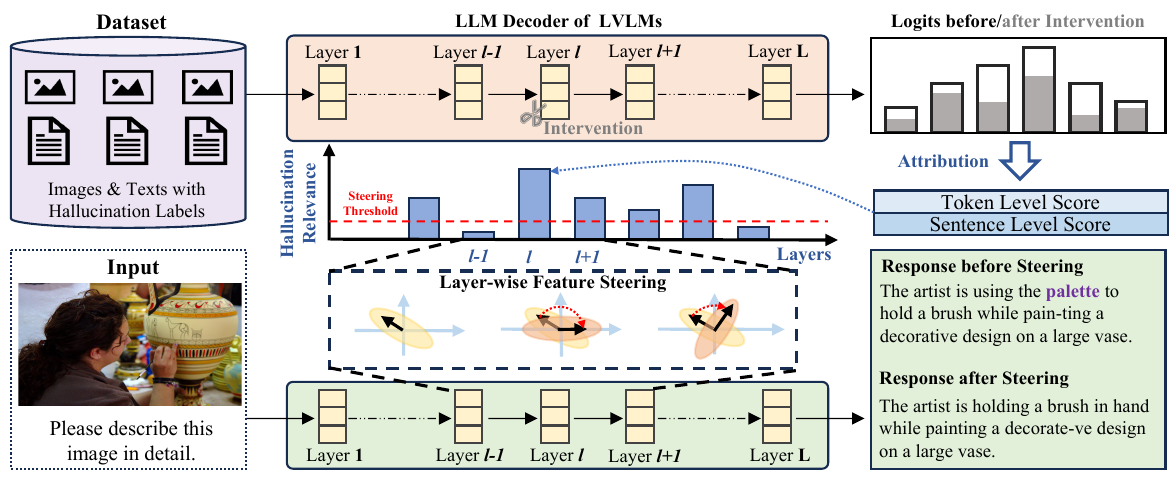} 
    \vspace{-2ex}
    \caption{Overview of our LTS-FS framework. First, we build a bi-granularity dataset with token-level and sentence-level hallucinations. Then, using the dataset, hallucination-relevant layers are identified via intervention-based attribution. Finally, a layerwise strategy is applied to control the feature steering intensity across layers according to the attribution scores.}
    \label{fig:Method}
    \vspace{-4ex}
\end{figure*}

To address these issues, we propose \textbf{L}ocate-\textbf{T}hen-\textbf{S}parsify for \textbf{F}eature \textbf{S}teering (\textbf{LTS-FS}), a plug-and-play framework that mitigates hallucinations while preserving the inherent capabilities of LVLMs (shown in Fig.~\ref{fig:Method}). 
First, to comprehensively quantify the contribution of the internal components of LVLMs to hallucinations, we construct a dataset with hallucination annotations in the generated text.
According to the prompt type and response length,
the hallucinations are categorized into two granularities, {\it i.e.}, the token-level and sentence-level. 
The former refers to cases where hallucinated content appears as specific tokens or short spans within a sentence, while the latter refers to cases where the entire natural sentence is hallucinated.
Our dataset is constructed based on binary QA~\cite{li2023evaluating} and open-ended generated captions~\cite{rohrbach2018object}.
Using the evaluation criteria of the benchmarks~\cite{rohrbach2018object,li2023evaluating,wu2025antidote}, the hallucinated content can be detected automatically, and then we categorize them into token-level or sentence-level by human experts.

Based on this dataset, we design an attribution method to quantify the contribution of each LVLM's layer to token-level hallucination. 
We further aggregate the token-level contributions to the sentence level and incorporate several weighted indicators inspired by prior analyses of hallucination patterns~\cite{huang2024opera,zhou2024analyzing}.
Our attribution method quantifies each layer's contribution to the logits of hallucinated outputs by masking the attention output of that layer and measuring the resulting change in hallucination-related outputs, yielding a quantitative attribution score. 
After conducting attribution across the full dataset, we average the layer-wise attribution scores from both token-level and sentence-level hallucination samples and use the aggregated scores to characterize each layer's contribution to hallucinations.

After obtaining layer-wise attribution scores, we propose a sparse layer selection for the feature steering method that maps the attribution scores to steering intensities, \textit{i.e.}, applying weaker steering to layers with low scores and stronger steering to those with high scores. 
By modifying only hallucination-relevant layers, we mitigate hallucinations while minimizing disturbance to the model's feature distribution, thereby better preserving its original capabilities.
Extensive experiments demonstrate that LTS-FS can further improve the hallucination mitigation performance of current advanced feature steering methods (\textit{e.g.}, a 2\% accuracy gain on POPE-popular with Qwen-VL2.5-7B) while better preserving generalization capability of LVLMs~(\textit{e.g.}, increasing detailness from 4.72 to 4.92 under GPT4v-Aided Evaluation on LLaVA-Bench). Our contribution can be summarized as follows: 

\begin{itemize}
    \item We categorize the hallucination into different granularities and construct a dataset to correlate model components with hallucinations at the token and sentence levels.
    \item We employ an intervention-based attribution method to locate hallucination-relevant layers by quantifying their contributions to hallucination outputs.
    \item We propose a layerwise sparse strategy that selectively adjusts feature steering intensity, achieving superior hallucination mitigation performance while preserving the generalization ability.
\end{itemize}

\section{Related Work}
\subsection{Hallucinations in LVLMs}
Hallucinations have been extensively studied in recent years~\cite{ji2023survey,huang2023survey,zhang2023siren,liu2024survey}.
Most methods are based on self-correction~\cite{yin2024woodpecker,bi2025asking}, instruction-tuning~\cite{wang2024vigc,liu2023mitigating,8957311}, or decoding-enhancement~\cite{huang2024opera,leng2024mitigating,An_2025_CVPR}. Typically, Yin {\it et al.}~\cite{yin2024woodpecker} refine textual responses while correcting hallucinations. Liu {\it et al.}~\cite{liu2023mitigating} compose negative instances to refrain from over-confidence. Huang {\it et al.}~\cite{huang2024opera} penalize specific tokens during decoding, which suppresses the formation of hallucinations. These methods generally require a large amount of manually labeled data and computing resources or suffer longer inference times.
To avoid these limitations, recent studies have proposed feature steering methods~\cite {yang2025nullu,liu2025reducing,li2023inference}. Yang {\it et al.}~\cite{yang2025nullu} project generated captions into a dedicated space to suppress hallucinated entities. Liu {\it et al.}~\cite{liu2025reducing} propose an intervention-based approach, steering the latent representations during inference with a pre-computed ``anti-hallucination" direction. However, directly adjusting weights or features may undermine internal knowledge and reduce generalization ability. To overcome these limitations, our method identifies hallucination-relevant layers and selectively adjusts features within them, thereby better preserving the internal knowledge of LVLMs.

\subsection{Parameter Localization}
Parameter localization, a technique that identifies parameters correlated with specific datasets, offers flexible and effective solutions for downstream tasks such as model fine-tuning~\cite{shen2024expanding}, knowledge editing~\cite{liu2025edit}, and model compression~\cite{wang2024rl}.
According to localization granularity, existing localization methods can be categorized into weight-level~\cite{hu2016network} and layer-level~\cite{dong2024prompt} paradigms.
For the weight-level paradigm, current methods design specific rules such as activations~\cite{hu2016network}, redundancy~\cite{srinivas2015data}, second derivatives~\cite{dong2017learning}, and energy efficiency~\cite{yang2017designing} to locate the data-relevant weights. 
For the layer-level paradigm, GRIFFIN~\cite{dong2024prompt} selects layers based on their high activation magnitudes in response to input prompts. FLAP~\cite{an2024fluctuation} computes the sample variance of each input feature as importance and locates layers accordingly. RL-Pruner~\cite{wang2024rl} determines the layer-wise importance distribution through reinforcement learning. 
Unlike the above methods designed for pruning or adjusting model parameters, we employ a layer-level strategy to locate the layers relevant to the hallucination phenomenon in LVLMs. The localization results can effectively support the feature steering process to mitigate hallucinations.

\subsection{Sparse Adjustments for Pre-trained Models}
To enhance the model capability in a specific domain while minimizing unintended disruptions to the overall model behavior, researchers have proposed sparse adjustment methods~\cite{liu2025edit, jia2024modelsparsitysimplifymachine, liu2024devilneuronsinterpretingmitigating, lin2025continuallearningsparsememory} that selectively modify a subset of model components. NMKE~\cite{liu2025edit} sparsely updates hidden neurons to edit the internal knowledge in LLMs. Jia {\it et al.}~\cite{jia2024modelsparsitysimplifymachine} develop a sparsity-aware method for model unlearning. BNS~\cite{liu2024devilneuronsinterpretingmitigating} selectively suppresses neuron activations to mitigate the social bias in pre-trained language models. Their sparse selection strategies are typically neuron-wise and designed for specific parameter adjustment methods~\cite{liu2025edit}. In contrast, we propose a layer-wise sparse selection strategy to enhance the feature steering paradigm for hallucination mitigation. This strategy is decoupled from any particular steering method, delivering consistently improved performance across different steering methods.

\section{Method}
In this section, we first construct the bi-granularity hallucination dataset~(\cref{Section:Bi-granularity Dataset Construction}). Based on this dataset, we introduce causal attribution to locate hallucination-relevant layers~(\cref{Section:Mechanism Level Localization}) and employ a layerwise sparse selection scheme to mitigate hallucination while maintaining the generalization ability of LVLMs~(\cref{Section:Layerwise Sparse Selection}).

\subsection{Bi-granularity Dataset Construction} \label{Section:Bi-granularity Dataset Construction}
To locate hallucination-relevant layers, we build a bi-granularity dataset by generating hallucination samples at the token and sentence levels, based on text length. 
Specifically, for single-sentence texts, their hallucinations can be annotated at the token level based on existing hallucination benchmarks~\cite{li2023evaluating,wu2025antidote}.
However, for multiple-sentence texts, token-level annotation is insufficient. 
As the length of generated text increases, the model's behavior evolves from producing isolated hallucinatory tokens to generating entire hallucinatory sentences~(\textit{i.e.}, removing them can significantly enhance the factuality with minimal impact on generation quality)~\cite{huang2024opera}.
Therefore, we categorize these samples as the sentence level for comprehensive localization of hallucination-relevant layers.

The bi-granularity hallucination samples are constructed based on current hallucination benchmarks such as CHAIR~\cite{rohrbach2018object}, POPE~\cite{li2023evaluating}, and Antidote~\cite{wu2025antidote}. 
Token-level samples are typically constructed by prompts phrased as wh-questions or yes/no questions. POPE and Antidote contain this type of question. 
Hallucination tokens can be identified by rule. 
For sentence-level hallucinations, we split multi-sentence texts and assess the image-grounded consistency of each sentence based on CHAIR, which is effective in identifying hallucination tokens. Sentences containing such tokens are labeled as hallucinatory.

\begin{figure}[t]
    \centering
    \includegraphics[width=0.95\linewidth]{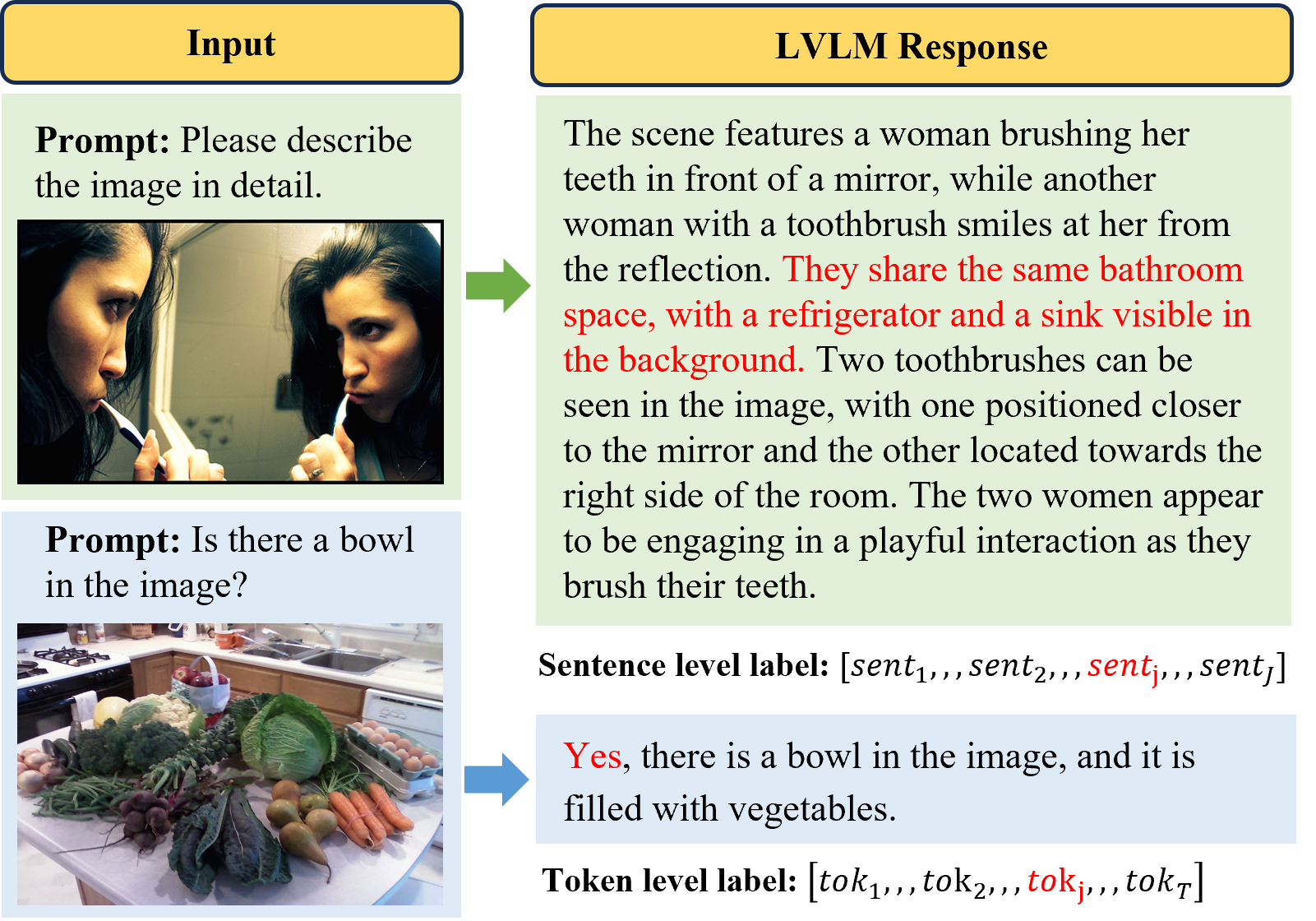} 
    \caption{Hallucination examples at token level and sentence level.}
    \label{fig:M31}
    \vspace{-4ex}
\end{figure}

\noindent\textbf{Examples of Both Granularities}.
At the token level, as shown in the upper sample of \cref{fig:M31}, the model generates a short response to a specific interrogative about a given item. In such cases, not all tokens are hallucinatory. Only ``the palette'' is absent from the image, while the remaining tokens describe objectively present content. 
At the sentence level, for longer and free-form responses in the lower sample of \cref{fig:M31}, the red part of the text reflects content conjectured from prior text and the image.
The entire clause following ``{reflection}'' is unsupported. 

\noindent\textbf{Data Usage and Split Protocol.} Note that all samples used to locate hallucination-relevant layers are computed only on the training split~(or a small calibration subset drawn from it) and do not include any samples from the evaluation benchmarks. 
Once the dataset is constructed, the subsequent policy is fixed and consistently applied to all following test evaluations without modification.

\subsection{Hallucination-Relevant Layer Localization}\label{Section:Mechanism Level Localization}
After constructing a dataset of images and texts annotated with hallucination labels, we use it to identify LVLM layers that are more prone to inducing hallucination~(\textit{i.e.}, hallucination-relevant layers).
Inspired by prior studies~\cite{yu2024neuron,yu2024understanding}, we design an attribution method that estimates the relevance between the hidden layers of LVLMs and the logits of hallucination outputs via causal intervention.

\noindent\textbf{Feed-Forward Process in LVLM Layers.}
Consider an LVLM composed of an image encoder, a projection module, 
and an LLM with $L$ layers. In the LLM decoding process, the output feature  $\mathbf{h}_l$ of layer $l$ is calculated as follows:
\begin{equation}\label{eq:llm-block}
\mathbf{a}_l=\texttt{MultiHeadAttn}(\mathbf{h}_{l-1}, h),
\end{equation}
\begin{equation}
\mathbf{x}_l = \texttt{MLP}\big(\texttt{LN}(\mathbf{h}_{l-1}+\mathbf{a}_l)\big),
\end{equation}
\begin{equation}
\mathbf{h}_l = \texttt{LN}(\mathbf{x}_l+\texttt{LN}(\mathbf{h}_{l-1}+\mathbf{a}_l)).
\end{equation}
Here, $\mathbf{a}_l$ and $\mathbf{x}_l$ are the outputs of the multi-head attention~(MHA) and MLP, respectively. $\texttt{LN}$ denotes the LayerNorm module. 
The MHA output $\mathbf{a}_l$ concatenates the output of $H$ heads. 
Given the output feature $\mathbf{h}_{l-1}$ at layer $l-1$, the attention output $\mathbf{a}_l$ and the parameters in subsquent layers $\theta_{\ge l}$, the logits of token $y$ is predicted 
as follows:
\begin{equation}
y \sim P_{\theta_{\ge l}}\big(y \mid \mathbf{h}_{l-1}, \mathbf{a}_l \big).
\label{eq:autoregression}
\end{equation}

\noindent\textbf{Layer-wise Attribution.}
To locate hallucination-relevant layers, we measure their contributions to hallucination outputs by introducing token-level and sentence-level attribution scores based on causal intervention techniques. 
Given the MHA output $\mathbf{a}_{l}$ of layer $l$,
and the output feature of the prior layer $\mathbf{h}_{l-1}$. The
attribution score at the token level $s_{\text{tok}}^{l}$ of layer $l$ is calculated as:
\begin{equation}
\label{eq:layer-score}
s_{\text{tok}}^{l} = \sum_{h=1}^H \log \left(\frac{P_{\theta_{\ge l}} (y \,\big|\mathbf{h}_{l-1}, \mathbf{a}_{l})}{P_{\theta_{\ge l}}(y \,\big|\mathbf{h}_{l-1}, \mathbf{a}_{l}\odot M^h)}\right).
\end{equation}
$M^h$ denotes a mask that sets the output of the $h$-th attention head to zero. We independently intervene on attention heads to measure the relevance between layers and hallucinatory tokens, building on prior studies~\cite{yu2024understanding,zhou2024analyzing} that such interventions enable more accurate estimation of how individual layers contribute to the logits of output tokens.

For the sentence level, we compute the attribution scores across all tokens in the sentence and aggregate them to obtain an overall attribution score, as the entire sentence is intrinsically associated with the hallucinated content~\cite{huang2024opera,zhou2024analyzing}. 
Since individual tokens vary in their contribution to hallucinations, we employ a weight-based aggregation method that assigns token weights based on several indicators, designed with insights from prior studies~\cite{huang2024opera,zhou2024analyzing}.
These studies suggest that (1) initial summarizing cues (\eg, additional) or the terminal punctuation of the preceding sentence, and (2) later tokens in the sentence, are more likely to trigger hallucination. Furthermore, (3) tokens exhibiting factual errors should also be emphasized. Therefore, we design three indicators to assign these tokens with higher weights. 
Given the set of tokens $T_{sent}$ in a sentence, the indicators for a token $y_t\in T_{sent}$ are defined as follows:\\
\textbf{(1) Cue indicator}: $u(y_t|T_{sent}) \in \{0,1\}$, where $u(y_t\mid T_{sent})=1$ if $y_t$ is a {summary token} (\eg, ``additional'' or a period); otherwise $u(y_t\mid T_{sent})=0$.\\
\textbf{(2) Position indicator}: $r(y_t|T_{sent}) \in [0,1]$, a higher value indicates later positions in the sentence.\\
\textbf{(3) Hallucination indicator}: $v(y_t|T_{sent})\in\{0,1\}$, where $v(y_t|T_{sent})=1$ if the token $y_t$ is identified as containing a factual error, and $v(y_t|T_{sent})=0$ otherwise.\\
A multiplicative weight is formed and then normalized:
\begin{equation}\label{eq:token-weights}
\begin{aligned}
\tilde w(y_t\mid T_{sent}) 
& = (1+\lambda_{\mathrm{cue}}\, u(y_t\mid T_{sent}))\\
& \times (1+\lambda_{\mathrm{pos}}\, r(y_t\mid T_{sent})) \\
&\times (1+\lambda_{\mathrm{hall}}\, v(y_t\mid T_{sent})),\\
w(y_t\mid T_{sent}) 
&= \frac{\tilde w(y_t\mid T_{sent})}{\sum_{y_k\in T_{sent}} \tilde w(y_k\mid T_{sent})},
\end{aligned}
\end{equation}
where $\lambda_{\mathrm{pos}}, \lambda_{\mathrm{cue}}, \lambda_{\mathrm{hall}}\!\ge\!0$ are hyperparameters that control the strength of the three indicators.
The attribution score for sentence-level hallucinations at layer $l$ is computed as the weighted sum of the token-level attribution scores $s_{tok}^{l}$.
\begin{equation}\label{eq:sent-score}
s_{\text{sent}}^{\,l}=\sum_{y_{t}\in T_{sent}} w(y_t\mid T_{sent}) \cdot s_{tok}^{\,l}.
\end{equation}
In practice, attribution scores are utilized according to specific tasks. For simple tasks such as question answering, a token-level score is employed due to the conciseness of model outputs. In contrast, the sentence-level score is adopted in more general tasks such as image captioning.
\subsection{Layerwise Feature Steering}\label{Section:Layerwise Sparse Selection}
After locating hallucination-relevant layers with higher attribution scores, an intuitive approach is to apply feature steering exclusively to these layers. In contrast to existing feature steering methods that uniformly steer all layers, layer-wise steering enables more targeted hallucination mitigation while minimizing unnecessary interference with the LVLM's internal representations.

Specifically, we propose a layer-wise steering strategy that combines hard sparsification and soft weighting. 
For layers with extremely low attribution scores, steering features of these layers have minimal impact on mitigating the model's hallucinations while substantially impairing its generalization capability. Therefore, we exclude such layers from the steering process by employing a mask parameterized by a threshold $r_s$.

For layers with high attribution scores, we scale the steering intensity proportionally to their normalized attribution scores $\tilde{s}^l$~(\textit{i.e.}, features in higher-scoring layers are steered more strongly). 
The soft weighting achieves a more favorable balance between mitigating hallucinations and preserving the model's generalization capability.

The detailed implementation of our layer-wise steering strategy is presented in \cref{alg:}. Since we only adjust the steering intensity, our method can be seamlessly integrated into existing feature steering methods~\cite{liu2025reducing,yang2025nullu}, as all of them inherently require an explicit setting of steering intensity. 
Moreover, given a fixed pre-trained LVLM, the layer-wise intensity derived by our framework is generalizable across diverse steering methods, highlighting its broad applicability and strong reusability.

\begin{algorithm}[h]
\caption{Feature Steering at layer $l$}\label{alg:}
\begin{algorithmic}[1] 
\REQUIRE layerwise attribution score $s^l$, mask threshold $r_s$, initial feature steering intensity $\lambda$, output features $\mathbf{h}_{l}$, and feature steering function $f:\mathbb{R}^d\times\mathbb{R}\rightarrow\mathbb{R}^d$.
\ENSURE steered feature $\tilde{\mathbf{h}}_{l}$
\STATE Computing threshold: $\tau = r_{s} \cdot  \frac{1}{L}\sum_{l=1}^{L} s^{l}$
\STATE Constructing mask: $m_l = \mathbf{1}\!\left[s^{\,l}\ge \tau\right]$
\STATE Masking attribution score $s^l = m_l  \cdot s^{l}$
\STATE Normalizing attribution score $\tilde{s}^l = \frac{s^l}{\sum_{i=1}^{L} s^i}$
\STATE Scaling feature steering intensity $\lambda_{l} = \lambda * m_{l} + \lambda \cdot \tilde{s}_{l} $
\STATE Steering feature $\tilde{\mathbf{h}}_l=f(\mathbf{h}_l, \lambda_l)$
\RETURN $\tilde{\mathbf{h}}_l$
\end{algorithmic}
\end{algorithm}
\vspace{-4ex}
\section{Experiments and Analysis}
In this section, we empirically investigate the effectiveness of LTS-FS in mitigating hallucinations while preserving model generation quality. 
Remarkably, we use 100 sentence-level hallucination samples and 100 token-level hallucination samples to construct the bi-level hallucination dataset for layer-wise attribution.
The sentence-level hallucination samples are selected and processed from CHAIR benchmark~\cite{rohrbach2018object}, while the token-level hallucination samples are from POPE~\cite{li2023evaluating} and Antidote~\cite{wu2025antidote}.
For more details about dataset construction, please refer to the supplementary material.

\begin{table*}[t]
\vspace{-1ex}
\caption{CHAIR results of various LVLMs on MSCOCO. \textbf{Bold} indicates the best performance. C\textsubscript{S} and C\textsubscript{I} mean lower hallucination. Recall and output length (Len.) serve as controls, indicating that reductions in C\textsubscript{S}/C\textsubscript{I} do not stem from suppressing objects or truncating responses. $ ^{*}$ denotes the feature steering methods.
}
\vspace{-2ex}
\label{tab:chair-eval}
\centering
\setlength{\tabcolsep}{6pt} 
\resizebox{0.9\textwidth}{!}{%
\begin{tabular}{l|cccc|cccc|cccc}
\toprule
\multirow{2}{*}{\textbf{Method}} & \multicolumn{4}{|c|}{\textbf{LLaVA-v1.5-7B}} & \multicolumn{4}{|c|}{\textbf{LLaVA-v1.5-13B}} & \multicolumn{4}{|c}{\textbf{Qwen-VL2.5-7B}} \\
& \textbf{C\textsubscript{S}}$\downarrow$  & \textbf{C\textsubscript{I}}$\downarrow$  & \textbf{Recall} & \textbf{Len.} 
& \textbf{C\textsubscript{S}}$\downarrow$  & \textbf{C\textsubscript{I}}$\downarrow$  & \textbf{Recall} & \textbf{Len.} 
& \textbf{C\textsubscript{S}}$\downarrow$  & \textbf{C\textsubscript{I}}$\downarrow$  & \textbf{Recall} & \textbf{Len.} \\
\midrule

Regular         &53.0 	&13.9 &	77.2& 	98.0 
                & 40.8 & 9.5& 	77.2 &  111.8 	  
                & 27.0 & 7.4  & 61.6  &120.6 \\  			 
VCD             &55.2 	&16.7 	&77.5 	&89.2 
                &39.2 & 9.2& 79.1& 		108.2  	 
                & 26.2 & 7.6&	61.2 &	120.3 \\ 	  
AGLA            & 50.8 & 16.1 & 75.2 &88.1
                & 38.4 & 9.1 & 78.7& 109.3 		 	
                &  25.2  & 7.1  &59.5 &118.6 \\		 	
$\text{Nullu}^{*}$           &50.2 & 13.7& 76.9& 93.3
                &38.0 & 9.4& 74.5 & 105.8	 		  
                & 27.4 &7.7 &60.7&121.6  \\	 	 	
$\text{VTI}^{*}$            & 47.4&13.9  &76.2 & 88.9   			
                & 36.3&9.2 &75.9 &  94.4 	 	 	
                & 25.5& 7.1& 61.6 & 121.3 \\  	 		
\midrule
\textbf{LTS-FS~(Nullu)}  & 46.8 & 13.5  &	76.6  & 	93.2
                & 35.7 & 8.9& 76.1 &  109.8	 
                &\textbf{23.8}  & 	\textbf{6.0} &60.8 &120.6  \\	 	
\textbf{LTS-FS~(VTI)}  & \textbf{35.8}  &  	\textbf{11.9} & 	75.4 & 	82.2
                & \textbf{32.0}& \textbf{8.8} & 74.2& 83.6 		 	
                & 24.8& 6.6& 62.5& 120.0\\ 			
\bottomrule
\end{tabular}
}
\vspace{-2ex}
\end{table*}

\begin{table*}[t]
\caption{Comparison of the average accuracy under different settings (i.e., Random, Popular, Adversarial) with different
baselines and our framework on POPE \textbf{Bold} indicates the best results, and \underline{underline} means the second best. $^{*}$ denotes the feature steering methods.
}
\vspace{-2ex}
\label{tab:pope-eval-acc}
\centering

\setlength{\tabcolsep}{5.5pt} 
\resizebox{0.95\textwidth}{!}{%
\begin{tabular}{l|ccc|ccc|ccc}
\toprule
\multirow{2}{*}{\textbf{Method}} & \multicolumn{3}{|c|}{\textbf{LLaVA-v1.5-7B}} & \multicolumn{3}{|c|}{\textbf{LLaVA-v1.5-13B}} & \multicolumn{3}{|c}{\textbf{Qwen-VL2.5-7B}} \\
& \textbf{Popular}  & \textbf{Random}  &  \textbf{Adversarial}
& \textbf{Popular}  & \textbf{Random}  &  \textbf{Adversarial}
& \textbf{Popular}  & \textbf{Random}  &  \textbf{Adversarial}   \\
\midrule

Regular      & 77.52& 85.37& 70.13	
             &78.40 & 81.91&71.07	
             &83.31 & 85.32& 80.17\\
VCD          & 79.09& 86.55&71.48
             & 79.38&82.27 &71.73
             & 83.19& 85.94&80.56\\
AGLA         &78.67 &85.32 &71.63
             &80.11 & 82.64&72.27
             &83.34 &86.02 &\underline{80.92}\\

$\text{Nullu}^{*}$&79.42 & 86.35& 71.57       
                  & 80.88 & 83.24& 72.43
                 & 83.06&85.82 &80.74\\
            
$\text{VTI}^{*}$  &77.03 & 84.84&69.40
                  & 79.22& 84.08&71.77
                 &82.74 &85.49 & 80.19\\
\midrule
\textbf{LTS-FS(Nullu)} & \textbf{80.09}& \textbf{87.13}&\underline{72.62}
                        & \underline{81.46} &\underline{83.96}&\underline{73.06}
                        & \textbf{83.59} & \textbf{86.21}&\textbf{81.11} \\
\textbf{LTS-FS(VTI)}& \underline{79.96}& \underline{86.77}&\textbf{73.04}
                 &\textbf{81.77}& \textbf{86.59}&\textbf{73.78}
                 &\underline{83.35}&\underline{86.04} &\underline{80.92} \\

\bottomrule
\end{tabular}
}
\vspace{-2ex}
\end{table*}

\begin{table*}[t]
\caption{Comparison of the average F1 score under different settings (i.e., Random, Popular, Adversarial) with different baselines and our framework on POPE. \textbf{Bold} indicates the best results, and \underline{underline} is the second best. $^{*}$ denotes the feature steering methods.
}
\label{tab:pope-eval-f1}
\centering
\setlength{\tabcolsep}{5.5pt} 
\resizebox{0.95\textwidth}{!}{%
\begin{tabular}{l|ccc|ccc|ccc}
\toprule
\multirow{2}{*}{\textbf{Method}} & \multicolumn{3}{|c|}{\textbf{LLaVA-v1.5-7B}} & \multicolumn{3}{|c|}{\textbf{LLaVA-v1.5-13B}} & \multicolumn{3}{|c}{\textbf{Qwen-VL2.5-7B}} \\
& \textbf{Popular}  & \textbf{Random}  &  \textbf{Adversarial}
& \textbf{Popular}  & \textbf{Random}  &  \textbf{Adversarial}
& \textbf{Popular}  & \textbf{Random}  &  \textbf{Adversarial}   \\
\midrule

Regular         & 80.71& 86.47&	75.85	
                 & 81.30&83.85 &76.47	
                 & 81.68&83.54 &	78.93 \\
VCD             & 81.23& 87.16&	76.04
                 & 82.01&83.76 &	75.76
                 & \underline{81.95}&\underline{83.88} &\underline{79.51}	 \\
AGLA            &81.47 &86.77 &	75.89
                &82.32& 83.58&	75.48
                 &81.86 &83.63 &	79.14\\

$\text{Nullu}^{*}$ &81.67 &86.28 & 76.17          
                 & 82.97&84.73 & 77.04
                & 81.27&83.73 &  79.32\\
                
$\text{VTI}^{*}$ & 80.40& 86.08& 75.42
                    & 81.83 &83.82 &76.80
                &80.88 &83.37 & 78.70 \\
\midrule
\textbf{LTS-FS(Nullu)}   &\underline{82.20} &\textbf{87.64}&\underline{76.22}
                        &\underline{83.42}& \underline{85.56}& \underline{78.36}
                        &\textbf{82.55}&\textbf{84.31} & \textbf{79.83}  \\
\textbf{LTS-FS(VTI)}   &\textbf{82.25} &\underline{87.32} & \textbf{77.32}
                & \textbf{83.58} &\textbf{87.48} & \textbf{79.91}
                &  81.38&\underline{83.88} &79.46 \\

\bottomrule
\end{tabular}
}
\vspace{-2ex}
\end{table*}

\subsection{Benchmarks and Baselines}
\noindent\textbf{Benchmarks.}
Following prior work, we evaluate our LTS-FS on typical benchmarks CHAIR~\cite{rohrbach2018object} and POPE~\cite{li2023evaluating}. Each metric is averaged across three independent runs with distinct random seeds. To assess overall performance after feature steering, we further include experiments on MME~\cite{fu2025mme} and LLaVA-Bench~\cite{liu2024improved}.

(a) \textbf{CHAIR}: Caption Hallucination Assessment with Image Relevance~\cite{rohrbach2018object} is a widely used benchmark to evaluate object hallucination in image captioning.
It quantifies the degree of object hallucination by calculating the ratio of all mentioned objects in the generated text that are not in the ground truth object set. 
There are two assessment criteria. CHAIR\textsubscript{S} quantifies the degree of object hallucinations at the sentence-level, while CHAIR\textsubscript{I} focuses on the instance level.
Lower C\textsubscript{S} and C\textsubscript{I} indicate fewer hallucinations.
In addition, we also report Recall and Sentence Length to ensure a fair comparison, since the reported hallucination metrics may be affected by the amount of generated content.

(b) \textbf{POPE}: Polling-based Object Probing Evaluation~\cite{li2023evaluating} contains 27,000 question-answer pairs about objects in MSCOCO~\cite{lin2014microsoft}, A-OKVQA~\cite{schwenk2022okvqa}, and GQA~\cite{hudson2019gqa}. These question-answer pairs involve only yes/no questions and are evenly distributed among existing and absent objects. There are three negative sample settings in each dataset, {\it i.e.}, random, popular, and adversarial~\cite{li2023evaluating}. This benchmark is evaluated using classification metrics: Accuracy, Recall, Precision, and F1-Score.

(c) \textbf{MME}: Multi-modal Large
Language Model Evaluation~\cite{fu2025mme} is a comprehensive evaluation benchmark for LVLMs that assesses their perception and cognition abilities. It comprises 10 perception-related and 4 cognition-related tasks, evaluated using binary classification. 	
MME is employed to measure hallucination while also capturing aspects of general model ability.

(d) \textbf{LLaVA-Bench}: LLaVA-Bench \cite{liu2024improved} comprises 24 images, each accompanied by a detailed, manually crafted description and a set of meticulously selected questions. Although this collection is relatively small in scale, it poses greater challenges for LVLMs.
We use GPT-4v to evaluate the model’s generations, assessing general capability.

\noindent\textbf{Baselines.}
We integrate our framework with two feature steering methods, Nullu~\cite{yang2025nullu} and VTI~\cite{liu2025reducing}.
To validate the utility of our methods, we evaluate the effectiveness of these two models, LTS-FS~(Nullu) and LTS-FS~(VTI), on three mainstream large
vision-language models, including LLaVA-v1.5-7B~\cite{liu2023visual}, LLaVA-v1.5-13B~\cite{liu2023visual} and Qwen-VL2.5-7B~\cite{bai2025qwen2}.
We compare our method with state-of-the-art baselines: VCD~\cite{leng2024mitigating}, AGLA~\cite{An_2025_CVPR}, Nullu~\cite{yang2025nullu}, and VTI~\cite{liu2025reducing}.

\noindent\textbf{Implementation}: 
For other hallucination mitigation methods, we use the default settings.
In our methods, we set $\lambda_{\mathrm{pos}}=1$, $ \lambda_{\mathrm{cue}}=1$, $\lambda_{\mathrm{hall}}=1$, and $r_{s} = 0.5$.
More implementation details can be found in the supplementary material.

\subsection{Results on CHAIR}
In CHAIR evaluation, we use \textit{Please describe the image in detail.} as the prompt. The results shown in ~\cref{tab:chair-eval} confirm that LTS-FS consistently outperforms the evaluated methods. The lowest CHAIR\textsubscript{S} and CHAIR\textsubscript{I} indicate our framework can better integrate visual knowledge and effectively reduce hallucinations. Comparison with Nullu and VTI demonstrates that our strategy can further enhance the performance of feature-steering-based methods. Moreover, the Recall and Length of our method are comparable to those of other methods. This provides partial evidence that our method mitigates hallucinations without sacrificing generation quality. For the evaluation of text generation quality, please refer to the supplementary material.

\noindent\textbf{Time Analysis.} Compared with decoding-based methods~(VCD and AGLA), feature steering methods~(Nullu and VTI) do not involve time-consuming additional processes in inference; thus, the inference speed is similar to that of the regular setting. LTS-FS also inherits this favorable characteristic, as shown in the supplementary material.

\begin{figure}
    \centering
    \includegraphics[width=0.95\linewidth]{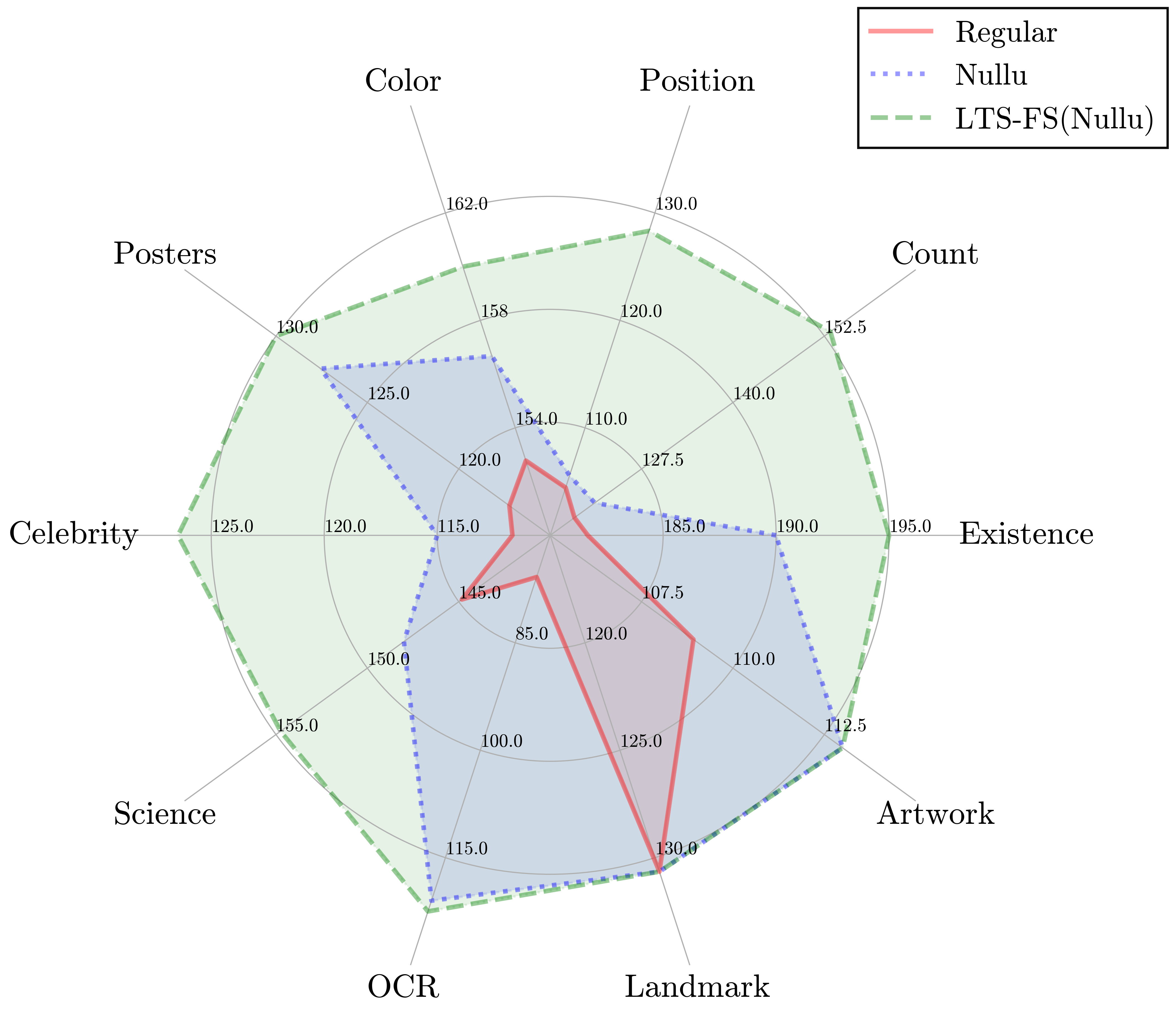}
    \caption{Results of MME evaluation.}
    \label{fig:MME results}
    \vspace{-2ex}
\end{figure}

\subsection{Results on POPE}
We conduct evaluations on POPE benchmark under the Popular, Random,
and Adversarial settings. Here, we mainly provide the average results of Accuracy and F1-score, respectively shown in~\cref{tab:pope-eval-acc} and~\cref{tab:pope-eval-f1}. The comprehensive results are available in the supplementary material. 
Since we use Qwen-VL for evaluation, some methods (\textit{e.g.}, Nullu) did not report corresponding results. Therefore, we reproduce all methods under an environment as consistent as possible to ensure fair comparison.

Experiments show that our method achieves the best accuracy and F1-score under all settings. Particularly, when using LLaVA-v1.5-13B, it increases the accuracy of the Random setting from 81.91\% to \textbf{86.59\%}. The results demonstrate the effectiveness of LTS-FS in mitigating hallucinations and its broad applicability across diverse open-source LVLMs and datasets.

\subsection{Results on MME}

We present the results of LLaVA-v1.5-7B on MME benchmark as a representative to evaluate the general ability of the model.
As shown in~\cref{fig:MME results}, LTS-FS consistently achieves improved performance across all perception-related tasks in MME. It is worth noting that Nullu achieves significant improvements mainly in tasks such as OCR and Posters, but has negligible effects in some tasks (e.g., Count). This likely indicates that typical feature-steering-based methods are susceptible to changes in feature distributions, whereas our layer-wise strategy can better ensure the model's comprehensive capability. More details are provided in the supplementary material.

\begin{table}[t]
\centering
\caption{Ablation results for different levels of hallucination.}
\vspace{-1ex}
\small
\renewcommand{\arraystretch}{1}
\begin{tabular}{c|cccc}
\toprule
Hallucination Level & C\textsubscript{S} & C\textsubscript{I} & POPE acc & POPE f1\\
\midrule
Baseline (Nullu)& 50.2 &	13.7 &	79.11	&81.37\\
Token-level only &50.0&	13.4&	79.59&	81.85\\
Sentence-level only &47.3&	\textbf{13.0}&	79.33	&81.58  \\
Both levels &\textbf{46.8} 	&13.5 	&\textbf{79.92}	&\textbf{82.02} \\
\hline
\end{tabular}
\label{tab:ablation-token-sentence-level}

\end{table}

\begin{table}[t]
\centering
\caption{Ablation results for different choices of $r_s$. }
\vspace{-1ex}
\small
\renewcommand{\arraystretch}{1}
\begin{tabular}{c|cccc}
\toprule
$ r_{s}$ & CHAIR\textsubscript{S} & CHAIR\textsubscript{I} & Recall & Length \\
\midrule
0.0 (Regular) & 53.0 	 & 13.9 	& 77.2 &	98.0 \\
0.3 &	49.5 & 14.2	& 76.2	 & 95.7 \\
0.5 & \textbf{46.8} 	 &\textbf{13.5} 	 & 76.6 	& 93.2\\
0.7 & 47.6 &  	13.0 	&76.6 	& 97.0 \\
0.9 & 49.1  &	13.3 & 	 75.7	 & 96.8 \\
Soft Gating & 46.7 & 13.4 & 76.1& 94.5 \\
\bottomrule
\end{tabular}
\label{tab:ablation-rs}
\vspace{-3ex}
\end{table}

\begin{figure*}[!ht]
    \centering
    \begin{subfigure}[t]{0.53\linewidth}
        \centering
        \includegraphics[width=\linewidth]{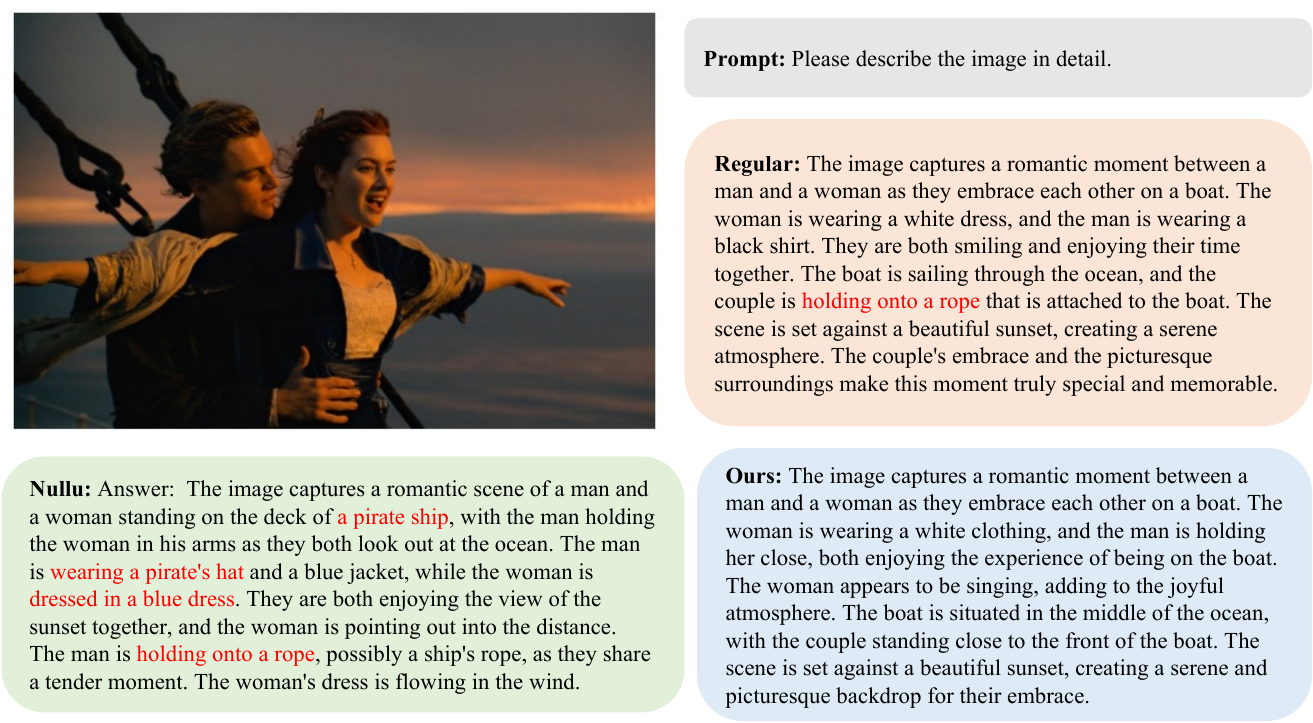} 
    \end{subfigure}
    \hfill
    \begin{subfigure}[t]{0.45\linewidth}
        \centering
        \includegraphics[width=\linewidth]{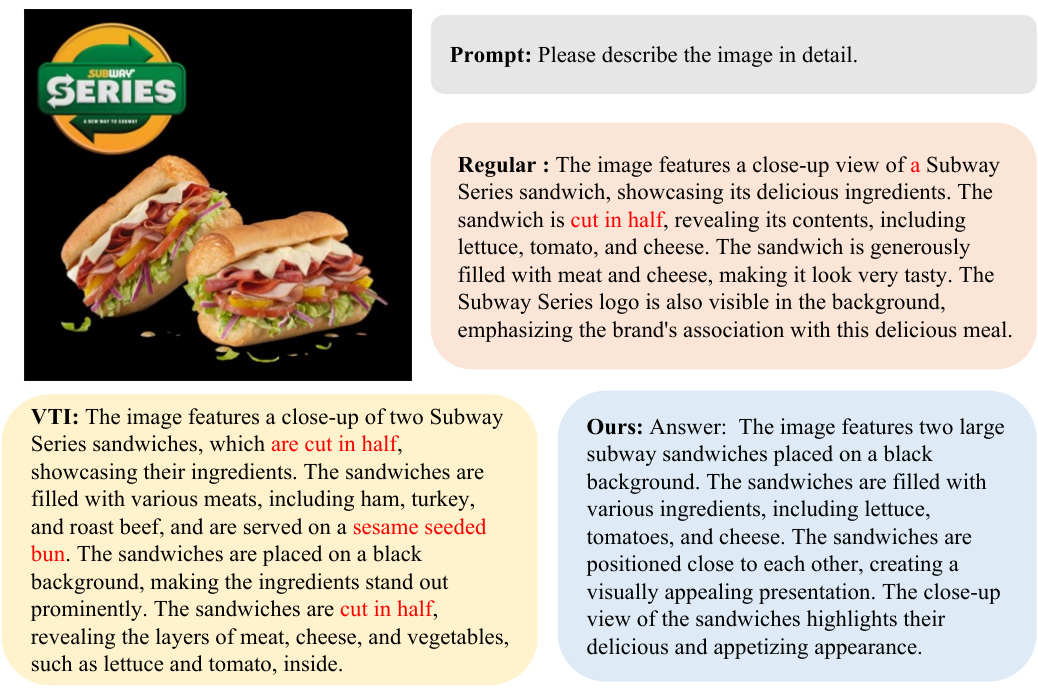}
    \end{subfigure}
    \vspace{-1ex}
    \caption{Demonstration of our framework for hallucination mitigation on two examples of LLaVA-Bench using LLaVA-v1.5-7B.}
    \label{fig:llava_bench}
    \vspace{-1ex}
\end{figure*}

\subsection{Ablation Studies}

\noindent\textbf{Effect of Two Levels of Hallucination}.
We discuss the effect of different levels of hallucination by respectively evaluating LTS-FS on the sub-datasets with only token-level hallucination and only sentence-level hallucination.
Results are shown in \cref{tab:ablation-token-sentence-level}, regarding the performance of Nullu as baseline. 
``Token-level only'' setup refers to calculating attribution scores based solely on token-level hallucination samples during the layer localization process.
In contrast, the ``Sentence-level only'' setup is only based on sentence-level hallucination samples. ``Both levels'' setup is equivalent to the overall framework of LTS-FS.

Results on the CHAIR benchmark indicate that layer localization based on sentence-level hallucinations achieves a more significant mitigation effect. This further demonstrates that sentence-level hallucination attribution is particularly beneficial for longer outputs, consistent with intuitive expectations. In contrast, layer localization based on token-level hallucinations is better suited to short responses in POPE. However, the ``Both level'' setup achieves the optimal performance in the POPE evaluation, which indicates that integrating sentence-level attribution is more conducive to enhancing the model's robustness. 
More detailed results are shown in the supplementary material.
These phenomena indicate that combining hallucination samples at the two hallucination levels can expand the conceptual range of hallucination attribution, thereby enabling more precise layer-wise localization.

\noindent\textbf{Selection of Mask Threshold $r_{s}$.}
The hyper-parameter $r_s$ directly determines how many layers should be steered. We compare results on the CHAIR across a set of candidate $r_{s}$ values to examine the impact of the mask threshold on generation performance. The LVLM is LLaVA-v1.5-7B, and the feature-steering basis method is Nullu. As shown in~\cref{tab:ablation-rs}, our strategy consistently improves hallucination mitigation. Meanwhile, the differences in results caused by $r_s\in[0.5,0.7]$ are negligible, which indicates that non-extreme selections of $r_s$ are sufficient to improve the performance of the feature steering method.

We also investigate a \emph{soft gating} variant for selecting $r_s$. Compared with fixing $r_s$ for all samples, soft gating sets $r_s$ per sample based on the attribution-score distribution across layers, allowing the number of steered layers to vary with the input.
As shown in~\cref{tab:ablation-rs}, soft gating performs comparably to hard gating with negligible differences, so we use hard gating with a fixed $r_s$ in all experiments for simplicity.

\begin{table}[t]
\centering
\caption{GPT-4V-aided evaluation of LLava-Bench.}
\vspace{-1ex}
\small
\renewcommand{\arraystretch}{1}
\begin{tabular}{l l c c}
\hline
Model & Method & Accuracy$\uparrow$ & Detailedness$\uparrow$ \\
\hline
\multirow{2}{*}{LLaVA-1.5} & Original & 5.74 & 5.23 \\
                           & Nullu    & 6.46 & 5.51 \\
                            & \textbf{LTS-FS(Nullu)} & \textbf{6.96}& \textbf{6.23}\\
\hline
\multirow{2}{*}{Qwen-vl2.5} & Original & 6.06 & 5.54 \\
                            & Nullu    & 6.37 & 5.68 \\
                            & \textbf{LTS-FS(Nullu)} & \textbf{6.59}& \textbf{6.07}\\
\hline
\end{tabular}
\label{tab:llava-in-the-wild-gpt4v}
\vspace{-5ex}
\end{table}

\subsection{Further Analysis}
\textbf{Case Study on LLaVA-Bench.} In~\cref{fig:llava_bench}, we provide two case studies based on LLaVA-v1.5-7B. More examples can be found in the supplementary material. 
The examples clearly show that hallucinations persist in typical feature-steering methods, where nonexistent details such as ``cut-in-half sandwiches'', ``pirate ships'', and ``pirate hats'' are fabricated. Our method consistently avoids these errors and produces descriptions that remain faithful to the visual content. These qualitative results demonstrate that our layer-wise feature steering effectively suppresses hallucinations while better preserving the model's comprehensive capabilities and fluent presentation.

\noindent\textbf{GPT-4V Aided Evaluation on LLaVA-Bench.}
Following Nullu~\cite{yang2025nullu}, we evaluate the performance of LTS-FS on LLaVA-Bench using GPT-4V Aided Evaluation. 
The evaluation prompt is shown in the supplementary material. 
The results are shown in~\cref{tab:llava-in-the-wild-gpt4v}, which demonstrate that LTS-FS can mitigate hallucination while better maintaining generation ability.

\section{Conclusion}
In this paper, we propose a plug-and-play framework LTS-FS, which mitigates hallucinations in LVLMs through feature steering while better preserving their generalization ability.
We first construct a bi-level hallucination dataset. 
Using this dataset, we attribute hallucination-relevant layers via causal interventions and assign a layer-wise score to each layer.
Finally, we design a layerwise strategy to selectively control steering intensity based on attribution scores across layers. 
Extensive experiments demonstrate that LTS-FS can effectively mitigate hallucinations while preserving the generalization ability of LVLMs. 
For future work, we will investigate the characteristics of hallucination-relevant layers detected by LTS-FS and aim to integrate the LTS-FS framework into the model pre-training process to reduce hallucination generation more fundamentally.

\noindent\textbf{Acknowledgements.}
This work was supported in part by the National Natural Science Foundation of China: 62236008, and in part by the Natural Science Foundation of Beijing under grant number L251082.
The authors would like to thank all the anonymous reviewers for their insightful comments. 

{
    \small
    \bibliographystyle{ieeenat_fullname}
    \bibliography{main}
}

\maketitlesupplementary

\section{Details of the construction of the dataset}\label{sec:details of construction}
In this section, we introduce the details of how to construct the Bi-granularity Dataset.

At first, to preserve generalization, the data used for dataset construction and the data used for experiments are strictly disjoint.
Particularly, for data selected based on CHAIR and POPE, we use data from the train split of MSCOCO. For data selected from Antidote, we do not use these data for evaluation.

\begin{figure*}[t]
    \centering
    \includegraphics[width=0.98\linewidth]{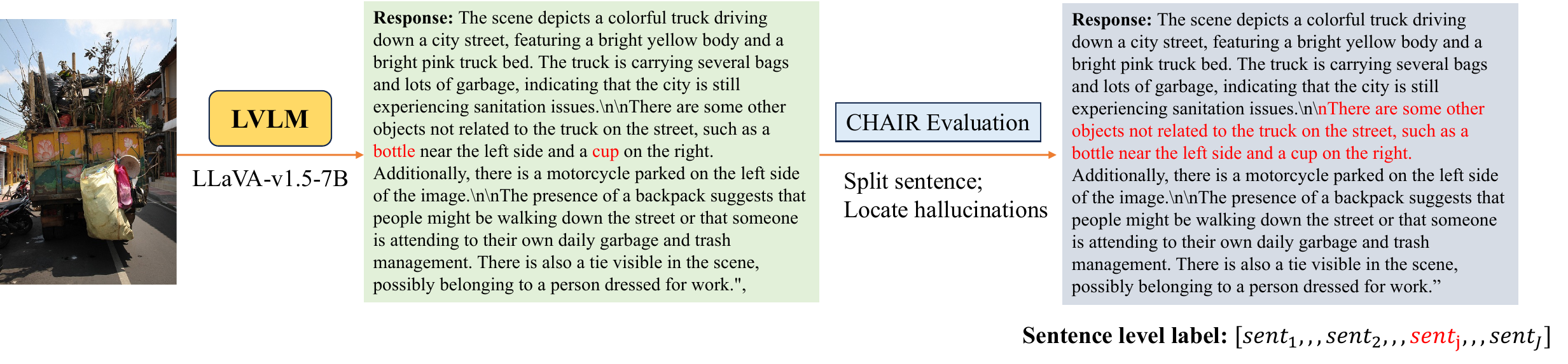}
    \vspace{-2ex}
    \caption{A sample generation based on CHAIR benchmark}
    \label{fig:construction_case}
\end{figure*}

Secondly, we explain how to get a single piece of data. As an example from CHAIR, the data instance is generated by an LVLM. We use LLaVA-v1.5-7B to produce a response according to the CHAIR benchmark, as illustrated in~\cref{fig:construction_case}. We then apply CHAIR’s evaluation criteria to detect hallucination and annotate the instance under our two-level scheme (token- and sentence-level). 
And then a piece of data is generated.
If the responses don't have hallucination, they are just not selected.

Finally, to balance two levels of data, we select 100 sentence-level samples and 100 token-level samples. All data are manually inspected to ensure accuracy.

\section{Implementation details of LTS-FS}\label{sec:details of lts-fs}
\noindent\textbf{Hyper-parameters.} 
The strength control parameters of $s^{l}_{tok}$: $\lambda_{cue},\lambda_{pos},\lambda_{hall}$ is set to be 1.
The mask threshold $r_{s}$ is selected to be 0.5, as shown in Tab.~\ref{tab:ablation-rs} of the main text.

\noindent\textbf{Environment.} All the experiments are conducted on one A100 80G. For the 7B model, two RTX3090 24G can replace an A100. For detailed Python requirements, please refer to our released code.

\section{Implementation Settings of CHAIR Results}\label{sec:details of chair}
\noindent\textbf{Generation Setting.} Here we set the generation config as follows: \textbf{Max\_New\_Tokens=128}, \textbf{num\_beams=1}, and \textbf{sampling=False}.

\noindent\textbf{Compared methods.} We employ the default parameters and settings as reported in the original papers.

\section{Generation Capability.}\label{sec:app-generation}
To evaluate general capability more comprehensively, we perform an evaluation using a broader benchmark called CLAIR~\cite{clair_2023}. This result in Tab.~\ref{tab:app-generation} shows that LTS-FS achieves a better trade-off between hallucination mitigation and general capability preservation.

\begin{table}[ht]
\centering
\scriptsize
\caption{Trade-off between hallucination mitigation and general capability preservation. }
\label{tab:app-generation}
\begin{tabular}{c|c|c|c|c}
\hline
\textbf{Method} & \textbf{CHAIRs} & \textbf{POPE acc} & \textbf{details} & \textbf{CLAIR} \\
\hline
Original & 53.0   & 77.63 & 5.23 & 80.03 \\

nullu & 50.2 & 79.11 & 5.51 & 75.00 \\

LTS-FS(nullu) & 46.8 & 79.92 & 6.23 & 82.74 \\

Soft Gating  & 46.7 & 79.5 & 6.26 & 83.64 \\
\hline
\end{tabular}
\end{table}
\section{More details of POPE results}\label{sec:details of pope}
\noindent\textbf{Generation Setting.} Here we set the generation config as follows: \textbf{Max\_New\_Tokens=16}, \textbf{num\_beams=1}, and \textbf{sampling=False}.

\noindent\textbf{Compared methods.} We employ the default parameters and settings as reported in the original papers.

\noindent\textbf{Total Results.} The total results is shown in~\cref{Tab:total_pope}. 
Across all settings, our LTS-FS framework achieves the best accuracy and F1, demonstrating consistent effectiveness in hallucination mitigation. Compared with the original feature-steering methods, applying LTS-FS consistently improves both VTI and Nullu on hallucination-related metrics.
Although LTS-FS and the other methods trade wins on recall, LTS-FS consistently maintains higher precision. Since, in hallucination evaluation, precision is more indicative of mitigation quality, this further supports the strong performance of our approach.

\section{More details of MME results} \label{sec:app-mme}
We report the MME numerical results in~\cref{tab:mme per results}. The numerical results demonstrate that LTS-FS can strongly increase the mitigation ability of feature steering methods. 
Specifically, across the subset most related to hallucination: Count, and Position, LTS-FS achieves great improvements, highlighting its effectiveness in enhancing feature-steering–based mitigation.

MME includes not only perception-related tasks but also recognition-related tasks. We report these results in~\cref{tab:mme-recog}.
Despite the sparsity selection emphasizes hallucination related cues rather than recognition factors, LTS-FS still produces improvements on recognition-related tasks.

\begin{table*}[ht]
\centering
\footnotesize
\caption{Results on all MME perception-related tasks.}
\label{tab:mme per results}
\begin{tabular}{l c c c c c c c c c c c}
\toprule
Method & \textit{Existence} & \textit{Count} & \textit{Position} & \textit{Color} & \textit{Posters} & \textit{Celebrity} & \textit{Scene} & \textit{Landmark} & \textit{Artwork} & \textit{OCR} & \textbf{Total} \\
\midrule

Regular &182	&118&	105&	151&118	&112	&145&131	&108	&78 &1248 \\
 Nullu &190	&122	&106	&157	&128	&118	&148	&130	&114	&121 & 1334\\
 LTS-FS(Nullu) &195	&153&	128&	157&	130&	127&	155&	131&	113&	123 & 1412\\
\bottomrule
\end{tabular}
\end{table*}

\begin{table*}[ht]
\caption{Results on all MME recognition-related tasks.}
\label{tab:mme-recog}
\centering
\small
\begin{tabular}{l l c c c c c}
\toprule
Model & Method &
\shortstack{Common Sense\\Reasoning} &
\shortstack{Numerical\\Calculation} &
\shortstack{Text\\Translation} &
\shortstack{Code\\Reasoning} &
\shortstack{Total} \\
\midrule
\multirow{3}{*}{LLaVA-1.5-7B} & Regular
&110	&50	&50	&71&281 \\
& Nullu
&113	&59	&75	&77& 324 \\
& LTS-FS + Nullu & 120	 &59 & 75& 80& 334 \\ 
\bottomrule
\end{tabular}
\end{table*}

\begin{table}[t]
\centering
\caption{Time analysis comparison of different hallucination mitigation strategies. VCD represents a decoding-based method. Nullu represents a feature-steering-based method.}
\begin{tabular}{lcc}
\toprule
\textbf{Method} & \textbf{Preparation Cost} & \textbf{Inference Cost} \\
\midrule
Regular & -- & 1.31s  \\
VCD & 0s & 3.14s \\
Nullu & 30mins & 1.37s  \\
Ours & 90mins & 1.34s  \\
\bottomrule
\end{tabular}
\label{tab:time analysis}
\end{table}

\begin{table}[h]
\centering
\caption{Ablation study of indicators. HI, CI, PI respectively indicate hallucination indicator, cue indicator, and position indicator.}
\renewcommand{\arraystretch}{1}
\begin{tabular}{c|cccc}
\toprule
Settings & C\textsubscript{S} & C\textsubscript{S} & Recall & Length \\
\midrule
Regular& 53.0 	 & 13.9 	& 77.2 &	98.0\\
w/o HI  & 52.0& 14.0& 76.9 & 97.4 \\
w/o CI & 48.2 & 13.6& 77.1& 95.7\\
w/o PI & 47.6& 13.7& 76.9&94.3\\
\midrule
LTS-FS(Nullu)&\textbf{46.8} & \textbf{13.5} & 76.6  &93.2\\
\bottomrule
\end{tabular}
\label{tab:ablation-indicator}
\vspace{-2ex}
\end{table}

\begin{table}[h]
\centering
\caption{Results of the generalization test of our framework. We use LLaVA-v1.5-7B to conduct this experiment. C\textsubscript{S} and C\textsubscript{I} is the CHAIR\textsubscript{S} and CHAIR\textsubscript{I} under CHAIR benchmark. ACC and F1 mean the accuracy and F1 score in the GQA subset on POPE.}
\renewcommand{\arraystretch}{1}
\begin{tabular}{c|cccc}
\toprule
Settings & C\textsubscript{S} & C\textsubscript{I} & Acc & F1 \\
\midrule
Regular& 53.0 	 & 13.9 	& 75.47 & 79.83	\\
MSCOCO→GQA & --- & --- & 77.31 & 79.57 \\
GQA→MSCOCO &49.5 &13.2 & --- & ---\\
Antidote→GQA & --- & --- & 77.28 & 80.12 \\
Antidote→MSCOCO & 49.8& 13.7& --- & ---\\
\midrule
LTS-FS(Nullu) & 46.8 & 13.5 & 77.15& 80.63\\
\bottomrule
\end{tabular}
\label{tab:generalization}
\vspace{-2ex}
\end{table}

\begin{figure*}[h]
    \centering
    \includegraphics[width=0.9\linewidth]{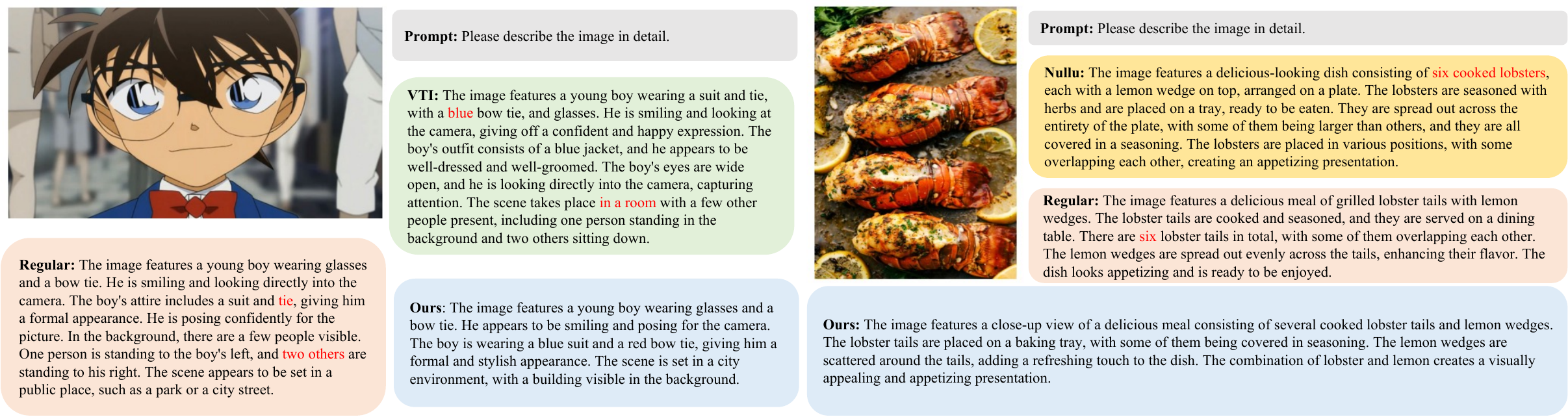}
    \caption{More examples on LLaVA-Bench.}
    \label{fig:supple_llava_bench}
\end{figure*}
\section{Time Analysis} \label{sec:app-time-cost}
There are two time cost analyses: the time to apply methods and the time for inference.
The time to apply methods is the time to employ a hallucination mitigation method in a specific LVLM. As an example, in order to apply VTI to LVLMs, the direction vector needs to be computed, and the layer should be adjusted. This whole time is the time to apply methods.
For our method, the time to apply the methods contains two parts. First, we need layer-wise attribution to select specific layers. Second, we need to apply feature steering methods based on these sparse layers. The second part time is almost the same as the original feature steering methods, which can be completed in under 30 minutes. The first part is the attribution process, which is time-consuming. For LLaVA-v1.5-7B, it takes about 1–2 hours on a single A100 80 GB GPU. 

As for the time for inference, our framework is based on feature steering methods. Therefore, the time for inference is comparable with regular generation. Comparison is shown in~\cref{tab:time analysis}. Despite requiring a longer preparation phase, the additional cost is reasonable, as it avoids the extra inference time latency that would otherwise accumulate during decoding and further highlights the inherent advantages of feature steering techniques.

\section{Ablation Study about Indicators}\label{sec:ablation of indicators}
In this section, we discuss the effect of the three indicators on sentence-level hallucination attribution. The result is shown in~\cref{tab:ablation-indicator}. We investigate the effect of removing each indicator in turn and find that \textit{w/o cue indicator} and \textit{w/o position indicator} yield only small changes, whereas \textit{w/o hallucination} causes a much larger decline, indicating that hallucination token attribution is paramount, with cue and position still providing auxiliary gains.

\section{Discussion about Generalization}\label{sec:discussion of generalization}
To assess generalization beyond the construction sources, we evaluate on datasets whose distributions differ from those used to build our bi-granularity labels. Although the construction leverages CHAIR, POPE, and Antidote, we additionally report results on MME and LLaVA-Bench, which serve as an out-of-distribution dataset of overall capability. We also run a decoupled calibration evaluation protocol: layer scores and weights are calibrated on one source (e.g., CHAIR on MSCOCO), then frozen and applied to a different target set for evaluation (e.g., POPE-GQA or Antidote). Concretely, CHAIR relies on MSCOCO; POPE uses MSCOCO and GQA; Antidote uses its own corpus. We therefore test cross-dataset pairs such as MSCOCO→GQA to verify transfer. The results are shown in~\cref{tab:generalization}).
MSCOCO→GQA denotes calibrating attribution on MSCOCO and evaluating on the POPE–GQA subset, and GQA→MSCOCO means attribution based on the GQA dataset and evaluation on the MSCOCO dataset under the CHAIR benchmark.
Despite calibrating on only part of the data, our framework typically delivers additional gains. The findings suggest that our improvements are driven by intrinsic generalization capacity, not by overfitting to a particular data distribution.

\section{More cases in LLaVA-bench}\label{sec:details of llava-bench}
More case studies on the LLaVA-bench are presented
in~\cref{fig:supple_llava_bench}, which demonstrates the effectiveness of our framework in hallucination mitigation. In particular, color and count attributes are given greater emphasis, thereby avoiding hallucinations in these aspects.

\section{GPT4v-Evaluation prompt}\label{sec:gpt4v prompt}
\begin{figure*}[h]
    \centering
    \includegraphics[width=0.9\linewidth]{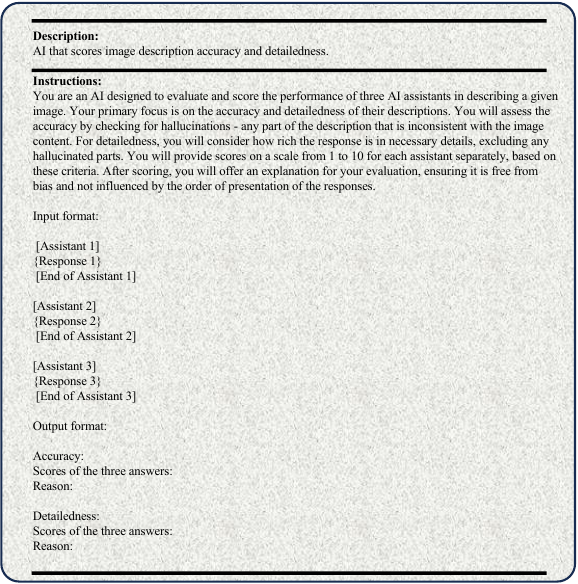}
    \caption{Prompt of GPT-4V Evaluation.}
    \label{fig:prompt}
\end{figure*}

Following VCD, the prompt for GPT4v-aided evaluation is shown in~\cref{fig:prompt}. The GPT4v receives three types of LVLM's responses and then generates output. Then we collect the output from GPT4v and finally report the average accuracy and detailedness.

\section{Limitation and future work}\label{sec:limitation}
Although our approach can be effectively ported to feature-steering methods and achieves strong hallucination mitigation, there is still room for development. Since existing feature steering techniques have not been evaluated on larger 70B-scale models, extending our method to 70B models remains a challenge.
We aim to extend our framework to larger models and further investigate its impact across additional multimodal domains.

\clearpage

\begin{table*}[h]
    \centering
    \caption{Average POPE results with Random and Popular.}
    \label{Tab:total_pope}
    \footnotesize
    \begin{tabular}{lcccccc}
    \toprule
    \textbf{Setting} & \textbf{Model} & \textbf{Method} & \textbf{Accuracy $\uparrow$} & \textbf{Precision$\uparrow$} & \textbf{Recall $\uparrow$} & \textbf{F1 Score $\uparrow$} \\
    \midrule
    \multirow{21}{*}{Random} & \multirow{7}{*}{LLaVA-v1.5-7B} & Regular & 85.37 & 80.77 & 93.22 & 86.47 \\
    & & VCD           & 86.55 & 84.02 & 90.69 & 87.16 \\
    & & AGLA          & 85.32 & 83.56 & 91.34 & 86.77 \\
    & & Nullu         & 86.35 & 84.36 & 91.09 & 86.28 \\
    & & VTI           & 84.84 & 80.02 & 93.36 & 86.08 \\
    & & LTS-FS(Nullu) & \textbf{87.13} & 84.69 & 91.02 & \textbf{87.64} \\
    & & LTS-FS(VTI)   & 86.77 & 84.13 & 91.00 & 87.32 \\
    \cline{2-7}
        & \multirow{7}{*}{LLaVA-v1.5-13B} & Regular        & 81.91 & 75.84 & 93.82 & 83.85 \\
    & & VCD            & 82.27 & 75.97 & 92.68 & 83.76 \\
    & & AGLA           & 82.64 & 76.19 & 93.16 & 83.58 \\
    & & Nullu          & 83.24 & 77.93 & 92.89 & 84.73 \\
    & & VTI            & 84.08 & 76.29 & 93.04 & 83.82 \\
    & & LTS-FS(Nullu)  & 83.96 & 78.89 & 93.85 & 85.56 \\
    & & LTS-FS(VTI)    & \textbf{86.59} & 82.35 & 93.47 & \textbf{87.48} \\
    \cline{2-7}
        & \multirow{7}{*}{Qwen-VL2.5-7B} & Regular        & 85.32 & 96.38 & 73.57 & 84.03 \\
    & & VCD            & 85.94 & 97.13 & 74.11 & 83.89 \\
    & & AGLA           & 86.02 & 96.56 & 73.65 & 83.63 \\
    & & Nullu          & 85.82 & 97.17 & 73.93 & 83.73 \\
    & & VTI            & 85.49 & 96.85 & 73.51 & 83.37 \\
    & & LTS-FS(Nullu)  & \textbf{86.21} & 97.09 & 74.78 & \textbf{84.31} \\
    & & LTS-FS(VTI)    & 86.04 & 97.23 & 73.64 & 83.87 \\	
    \midrule
    \multirow{21}{*}{Popular} 
    & \multirow{7}{*}{LLaVA-v1.5-7B} & Regular        & 77.52 & 71.45 & 93.22 & 80.71 \\
    & & VCD            & 79.09 & 73.21 & 92.17 & 81.23 \\
    & & AGLA           & 78.67 & 75.39 & 89.02 & 81.47 \\
    & & Nullu          & 79.42 & 74.45 & 91.04 & 81.67 \\
    & & VTI            & 77.03 & 70.90 & 93.36 & 80.40 \\
    & & LTS-FS(Nullu)  & \textbf{80.09} & 75.28 & 91.07 & 82.20 \\
    & & LTS-FS(VTI)    & 79.96 & 75.25 & 91.14 & \textbf{82.25} \\	
    \cline{2-7}
        & \multirow{7}{*}{LLaVA-v1.5-13B} & Regular        & 78.40 & 71.78 & 93.76 & 81.30 \\
    & & VCD            & 79.38 & 72.24 & 92.47 & 82.01 \\
    & & AGLA           & 80.11 & 72.88 & 92.16 & 82.32 \\
    & & Nullu          & 80.88 & 74.91 & 93.02 & 82.97 \\
    & & VTI            & 79.22 & 73.26 & 93.04 & 81.83 \\
    & & LTS-FS(Nullu)  & 81.46 & 75.62 & 92.93 & 83.42 \\
    & & LTS-FS(VTI)    & \textbf{81.77} & 75.58 & 93.47 & \textbf{83.58} \\
    \cline{2-7}
        & \multirow{7}{*}{Qwen-VL2.5-7B} & Regular        & 83.31 & 91.14 & 74.58 & 81.68 \\
    & & VCD            & 83.19 & 90.27 & 74.18 & 81.95 \\
    & & AGLA           & 83.34 & 90.69 & 74.53 & 81.86 \\
    & & Nullu          & 83.06 & 91.20 & 74.04 & 81.27 \\
    & & VTI            & 82.74 & 90.86 & 73.51 & 80.88 \\
    & & LTS-FS(Nullu)  & \textbf{83.59} & 91.12 & 76.18 & \textbf{82.55} \\
    & & LTS-FS(VTI)    & 83.35 & 90.96 & 73.64 & 81.38 \\	
    \midrule
    \multirow{21}{*}{Adversarial} 
    & \multirow{7}{*}{LLaVA-v1.5-7B} & Regular        & 70.13 & 64.14 & 93.22 & 75.85 \\
    & & VCD            & 71.48 & 66.28 & 89.62 & 76.04 \\
    & & AGLA           & 71.63 & 66.59 & 90.13 & 75.89 \\
    & & Nullu          & 71.57 & 66.06 & 90.53 & 76.17 \\
    & & VTI            & 69.40 & 63.46 & 93.36 & 75.42 \\
    & & LTS-FS(Nullu)  & 72.62 & 65.99 & 90.62 & 76.22 \\
    & & LTS-FS(VTI)    & \textbf{73.04} & 67.37 & 91.24 & \textbf{77.32} \\	
    \cline{2-7}
        & \multirow{7}{*}{LLaVA-v1.5-13B} & Regular        & 71.07 & 64.60 & 93.83 & 76.47 \\
    & & VCD            & 71.73 & 63.61 & 94.23 & 75.76 \\
    & & AGLA           & 72.27 & 64.14 & 93.56 & 75.48 \\
    & & Nullu          & 72.43 & 66.08 & 92.44 & 77.04 \\
    & & VTI            & 71.77 & 65.58 & 93.01 & 76.80 \\
    & & LTS-FS(Nullu)  & 73.06 & 67.01 & 92.96 & 78.36 \\
    & & LTS-FS(VTI)    & \textbf{73.78} & 67.51 & 93.47 & \textbf{79.91} \\	
    \cline{2-7}
        & \multirow{7}{*}{Qwen-VL2.5-7B} & Regular        & 80.17 & 85.21 & 73.64 & 78.93 \\
    & & VCD            & 80.56 & 85.31 & 75.07 & 79.51 \\
    & & AGLA           & 80.92 & 85.73 & 74.72 & 79.14 \\
    & & Nullu          & 80.74 & 86.32 & 74.24 & 79.32 \\
    & & VTI            & 80.19 & 85.75 & 73.51 & 78.70 \\
    & & LTS-FS(Nullu)  & \textbf{81.11} & 86.14 & 75.07 & \textbf{79.83} \\
    & & LTS-FS(VTI)    & 80.92 & 85.94 & 73.64 & 79.46 \\
    \bottomrule
    \end{tabular}
\end{table*}

\end{document}